\documentclass[10pt,twocolumn,letterpaper]{article}

\usepackage{cvpr}              % To produce the CAMERA-READY version
\usepackage[dvipsnames]{xcolor}

\usepackage{pifont}
\usepackage{array}
\newcolumntype{x}[1]{>{\centering\arraybackslash\hspace{0pt}}p{#1}}

\usepackage{multirow}
\usepackage{rotating}
\usepackage[table]{xcolor}

\usepackage{times}
\usepackage{epsfig}
\usepackage{graphicx}
\usepackage{amsmath}
\usepackage{amssymb}
\usepackage{booktabs}
\usepackage{scalerel}

\newcommand{\tit}[1]{\smallskip\noindent\textbf{#1.}}
\newcommand{\method}{{GazeD}\xspace}

\definecolor{darkgreen}{rgb}{0.16,0.70,0.40}
\definecolor{customgreen}{RGB}{0,176,80} % Custom green color
\definecolor{darkyellow}{rgb}{0.8, 0.6, 0.0}

\usepackage{cite}

\definecolor{cvprblue}{rgb}{0.21,0.49,0.74}
\usepackage[breaklinks,colorlinks,citecolor=cvprblue]{hyperref}

\def\paperID{*****} % *** Enter the Paper ID here
\def\confName{3DV\xspace}
\def\confYear{2026\xspace}

\title{\method: Context-Aware Diffusion for Accurate 3D Gaze Estimation}
\author{
Riccardo Catalini$^{1,}$\thanks{Corresponding author: riccardo.catalini@unimore.it}
\and
Davide Di Nucci$^{1}$
\and
Guido Borghi$^{1}$
\and
Davide Davoli$^{2,}$\thanks{Providing contracted services for Toyota Motor Europe}
\and
Lorenzo Garattoni$^{2}$
\and
Gianpiero Francesca$^{2}$
\and
Yuki Kawana$^{3}$
\and
Roberto Vezzani$^{1}$\\[3pt]
$^{1}$University of Modena and Reggio Emilia \quad
$^{2}$Toyota Motor Europe \quad
$^{3}$Woven by Toyota\\[3pt]
{\small \texttt{\{rcatalini, ddinucci, gborghi, rvezzani\}@unimore.it,}}\\
{\small \texttt{\{davide.davoli, lorenzo.garattoni, gianpiero.francesca\}@toyota-europe.com, yuki.kawana@woven.toyota}}
}
\begin{document}
\maketitle

\begin{abstract}
We introduce \method, a new 3D gaze estimation method that jointly provides 3D gaze and human pose from a single RGB image. 
Leveraging the ability of diffusion models to deal with uncertainty, it generates multiple plausible 3D gaze and pose hypotheses based on the 2D context information extracted from the input image. Specifically, we condition the denoising process on the 2D pose, the surroundings of the subject, and the context of the scene. 
With \method we also introduce a novel way of representing the 3D gaze by positioning it as an additional body joint at a fixed distance from the eyes. The rationale is that the gaze is usually closely related to the pose, and thus it can benefit from being jointly denoised during the diffusion process. 
Evaluations across three benchmark datasets demonstrate that \method achieves state-of-the-art performance in 3D gaze estimation, even surpassing methods that rely on temporal information. Project details will be available at \url{https://aimagelab.ing.unimore.it/go/gazed}
\end{abstract}

\section{Introduction}
\label{sec:intro}

The importance of 3D gaze estimation lies in its ability to unlock deeper insights into human attention~\cite{raptis2017using}, and cognition~\cite{eckstein2017beyond}, which are central to a wide range of applications such as human-computer interaction~\cite{sharma2013eye}, behavioral analysis~\cite{kerr2019eye}, and extended reality systems~\cite{sitzmann2018saliency}, surveillance~\cite{vural2009eye}, autonomous driving~\cite{pal2020looking}, and robotics~\cite{palinko2015eye}. 

\begin{figure}[t!]
    \centering
    \includegraphics[width=\linewidth, page=2]{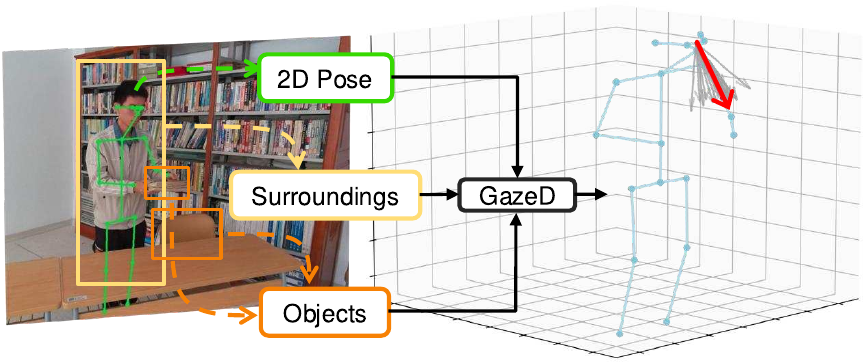}
    \caption{\method method jointly predicts 3D gaze and body pose analyzing the 2D pose, the surroundings of the subject and the context, in terms of objects in the scene.}
    \label{fig:first_image}
\end{figure}

Computer vision researchers have traditionally approached automated gaze analysis by dividing it into two main tasks~\cite{tonini2023object}: \textit{gaze estimation} and \textit{gaze target detection}, also referred to as \textit{gaze following}.
Specifically, gaze estimation aims to predict the direction of a person's gaze, while gaze target detection aims to pinpoint the exact location a person is looking at within the scene.

Methods for 3D gaze estimation are often based on the availability or extraction of detailed information about the human face or upper body~\cite{cheng2018appearance,funes2014eyediap,guo2020domain,kellnhofer2019gaze360}, ranging from the positions of the pupils to the exact location of the eyes. 
%However, these methods require careful annotations, as the use of automatic detectors instead leads to a notable performance drop~\cite{tu2022end}. 
Instead, only a few methods take advantage of the 
% information provided by the 
context, and even fewer works attempt to combine it with the human pose~\cite{toaiari2024upper}. These elements are used more often for gaze target detection~\cite{tonini2023object,guan2020enhanced}, as they are required to relate the target of the gaze to the elements in the scene.

However, we believe that the scene context and the human pose contain knowledge useful also for 3D gaze estimation: the context influences the gaze, and the gaze itself strictly depends on the body pose.
Indeed, previous studies have shown that gaze direction and body pose are closely interrelated \cite{Pose2Gaze}. Some works~\cite{toaiari2024upper,nonaka2022dynamic} utilize the 2D pose or the head or body orientation to estimate the 3D gaze. However, these methods lack a mechanism to directly correlate the pose with the final gaze output.

\begin{figure}[t!]
    \centering
    \includegraphics[width=.7\linewidth, page=5]{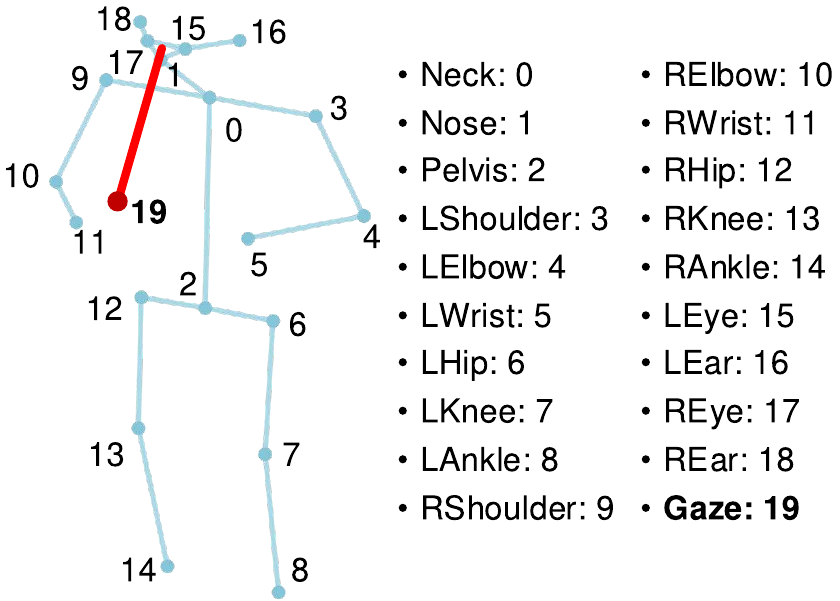}
    \caption{Human body skeleton with our additional gaze joint.}
    \label{fig:add_joint}
\end{figure}

Therefore, in this paper, we introduce \method, 
% a novel method 
that efficiently combines different elements from the scene, \ie the 2D body pose, the subject's surroundings, and the 
% information about the 
global context with objects, to output 
% a precise 
the 3D gaze direction (see Fig.~\ref{fig:first_image}).
A key idea of \method is to model the gaze as a virtual protrusion from the forehead of the person, between the eyes, where we placed \textbf{an additional joint} here referred to as \textbf{gaze joint} (see Fig.~\ref{fig:add_joint}).
The gaze joint has a variable direction (the gaze angle), while its distance from the head is fixed. Therefore, the gaze joint is not positioned in correspondence with the target object, as required to solve the gaze target detection task, but it acts as a proxy to compute the gaze direction. 

Having modeled the gaze direction using an additional joint enables the resolution of the problem as an extension of 3D human pose estimation. \method is thus based on a regression head, which outputs both pose and gaze working on a common embedding. 

Because of the intrinsic ambiguity of lifting 2D information to the 3D world, as well as the multiple possible gaze directions given the body posture and the context of the scene, \method regresses the 3D gaze and pose using a diffusion model. By conditioning the denoising process using 2D pose, surroundings, and context features, \method models the uncertainty in the data and generates multiple plausible output hypotheses. To our knowledge, we are the first to adopt a diffusion model to regress 3D gaze direction.

The embeddings used as conditioning for the diffusion model are generated by \method as two consecutive steps. 
Starting from the 2D pose estimated by an off-the-shelf method, the first step recovers information from the context close to the subject. The second step extracts additional cues from the objects in the scene, \ie it captures the context of the scene far from the subject. 

As an additional advantage, \method works on a single RGB image, avoiding the computational complexity of processing video sequences and the need for specific hardware. This streamlines its adoption in real-world applications, simplifies the training procedure, and facilitates the acquisition of new datasets. In contrast, 3D estimators based on sequences of frames~\cite{nonaka2022dynamic,guan2023end,jindal2024spatio} or specific data modalities, such as depth maps or point clouds~\cite{toaiari2024upper,hu2023gfie,funes2014eyediap}, have more limited applicability in real-world scenarios.

In summary, the contributions of our paper are: 
i) We introduce \method, a method for 3D gaze estimation that combines surroundings, context with objects, and 2D human pose features to condition a diffusion model. The denoising process produces multiple plausible hypotheses of 3D gaze and human pose.
ii) We propose a novel representation of the gaze as an additional joint;
% of the skeleton; 
as a consequence, the proposed method neatly outputs both the 3D gaze and the 3D human pose.
iii) Experimental evaluations on several datasets demonstrate that \method achieves 
state-of-the-art 
performance in 3D gaze estimation, even surpassing methods that use multiple input modalities. Additionally, our method also achieves high accuracy in predicting the 3D human pose.

\section{Related Work}

\subsection{3D Gaze Estimation} 
Recently, research in 3D gaze estimation has evolved significantly.
% over the past decades, with 
Approaches are broadly categorized into two categories~\cite{nonaka2022dynamic}: geometry-based and appearance-based ones.
% , and 2D-to-3D lifting. 

\noindent \textbf{Geometry-based methods.} 
These methods~\cite{lu2016estimating,hennessey2006single,lee20123d,zhu2005eye} rely on constructing a 3D model of the eye using optical or geometric properties. These techniques are accurate in controlled environments (\textit{e.g.} good light conditions~\cite{nakazawa2012point}) with consistent subject characteristics (\textit{e.g.} head position~\cite{valenti2011combining}). Unfortunately, they often require specialized and expensive hardware -- such as infrared cameras or eye-tracking devices -- extensive calibration, limiting their applicability in real-world settings. 

\noindent \textbf{Appearance-based methods.} 
These methods~\cite{fischer2018rt,zhang2020eth,zhang2017mpiigaze,cheng2018appearance} have gained popularity due to their reliance on standard RGB cameras to estimate the gaze from eye and face images directly. Early appearance-based approaches used hand-crafted features, such as pixel intensity or eye shape, but suffered from limited robustness in unconstrained environments. 
The advent of deep learning has significantly improved the performance of these methods, enabling more robust gaze estimation across varying lighting conditions, head poses, and subjects~\cite{cheng2024appearance}. 

Some methods~\cite{toaiari2024upper,hu2023gfie,funes2014eyediap,hu2022we} use RGB and depth data to recover scene depth, but this requires specialized hardware and is less suitable for outdoor use due to limitations like sunlight interference~\cite{sarbolandi2015kinect}. Alternatively, temporal information modeled with RNNs or LSTMs has improved gaze estimation by capturing movement patterns~\cite{nonaka2022dynamic,zhou2019learning,palmero2018recurrent}, though such approaches demand significant computational resources to handle long video sequences.

\begin{figure*}[th!]
    \centering
    \includegraphics[width=\linewidth, page=3]{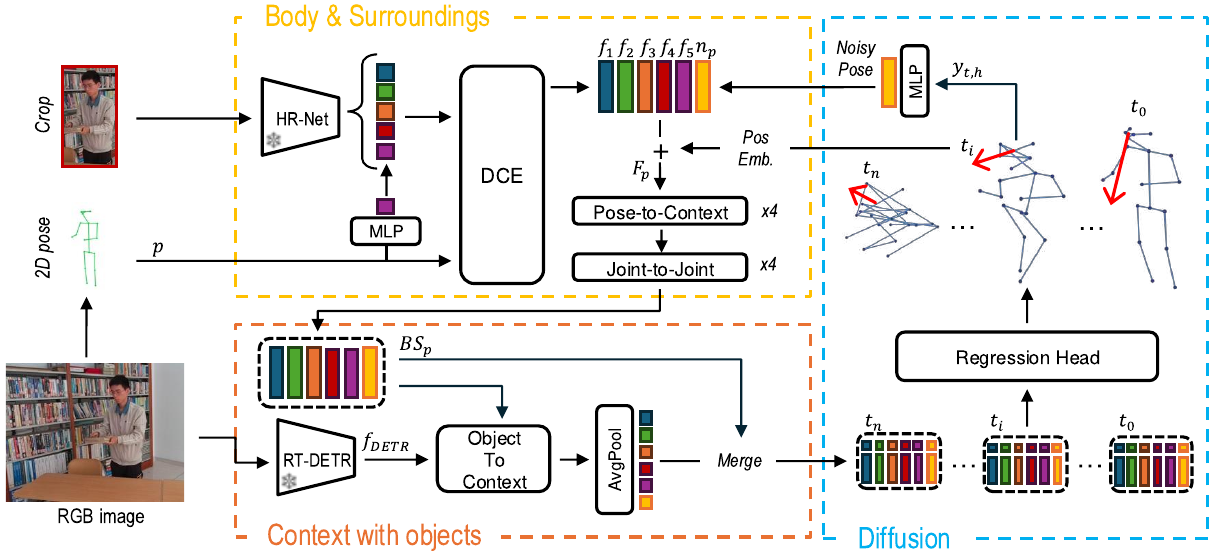}
    \caption{Overview of the proposed \method method that predicts the 3D gaze and human pose starting from a single input RGB image, combining information from the 2D body pose, surroundings, and context with objects.
    }
    \label{fig:general}
\end{figure*}
\subsection{3D Human Pose Estimation}

3D Human Pose Estimation (HPE) typically involves estimating 2D poses and then lifting them to 3D~\cite{d2021refinet,d2023depth}, a step that remains challenging due to the ambiguity of inferring 3D from 2D~\cite{cao2017realtime,chen2018cascaded}. To address this, some methods use temporal information~\cite{li2022mhformer,zheng20213d}, although this adds latency. Given the under-constrained nature of the problem~\cite{simo2012single}, multihypothesis approaches generate multiple plausible 3D poses instead of a single estimate, using techniques such as mixture density networks~\cite{oikarinen2021graphmdn} or conditional variational autoencoders~\cite{sharma2019monocular}.

\noindent \textbf{Diffusion Models.} 
More recently, Denoising Diffusion Probabilistic Models~\cite{ho2020denoising} have been applied to 3D HPE. These models treat 3D pose estimation as a reverse diffusion process, where a highly uncertain 3D pose distribution is progressively refined toward a more accurate pose. Methods like DiffPose~\cite{gong2023diffpose} leverage spatial-temporal context from 2D pose sequences to guide this diffusion process. A key advantage of diffusion models in this context is their ability to generate multiple hypotheses, providing a probabilistic framework naturally, and this allows for improved performance by aggregating multiple outputs, effectively reducing the impact of outliers. Furthermore, graph convolutional neural networks have been integrated with diffusion models~\cite{choi2023diffupose} to explicitly capture the correlations between joints, enhancing pose estimation accuracy.

\section{Method}
Given a single RGB image $I \in \mathbb{R}^{H \times W \times 3}$ as input, our goal is to predict the 3D gaze direction together with the 3D pose of the person in the scene. To this end, we define an additional gaze joint and we concatenate it to the list of body joints to provide the output $\textbf{y} \in \mathbb{R}^{J \times 3}$, where $J$ is the number of skeleton's joints, including the gaze joint (see Fig.~\ref{fig:add_joint}).
The gaze unit vector $v$ is defined as the direction from the midpoint between the eyes and the gaze joint:
\begin{equation}
    v =  \Big\Vert   \textbf{y}^{Gaze} - 
        \left(
            \frac{\textbf{y}^{LEye}+\textbf{y}^{REye}}{2} \right) 
    \Big\Vert
\end{equation}

% $\hat{ \textbf{y} } = \{p^{3d}_{j}\}_{j=0}^{J}$ where $p^{3d} \in \mathbb{R}^3$ and $J$ is the number of skeleton's joints. 

As shown in Figure~\ref{fig:general}, \method is composed of three modules: \textbf{Body \& Surroundings}, responsible for the feature extraction from the human pose and surroundings, \textbf{Context with objects} to integrate the general context, including the location of objects, and a \textbf{Diffusion} module that contains a regression head for the 3D gaze and pose prediction as well as the diffusion scheduler. 

\subsection{Body \& Surroundings}
% We assume to have an initial 2D pose $\textbf{p} \in \mathbb{R}^{J \times 2}$ pre-computed through a pose estimator as well as the corresponding crop of the image around the person. 
% Since the gaze joint is virtual and lacks a corresponding view in the scene, the midpoint between the eyes is used as a representative for the 2D pose. This midpoint is concatenated to the output of the 2D pose estimator to  obtain a final pose composed of $J$ joints.
This module takes as input a cropped image of the person, along with its 2D pose $\mathbf{p}\in \mathbb{R}^{J\times2}$. The pose $\mathbf{p}$ is obtained by concatenating the output of a 2D pose estimator with an additional joint representing the 2D gaze. However, as the gaze joint is virtual and lacks a correspondence in the image, we set the 2D gaze point as the midpoint between the eyes. %at a fixed distance from the face.
Although it is not the 2D projection of the corresponding 3D gaze joint, such a point has proven to be a good and consistent initialization. For this joint, it will not only be necessary to perform a third-dimensional lifting, but all three coordinates must be correctly estimated by the method.

We use a HR-Net\cite{sun2019deep} backbone to extract intermediate hierarchical features $\mathcal{H}=\{\mathbf{H}_l \in \mathbb{R}^{H_l \times W_l \times C_l}\}_{l=1}^L$, where $L$ is the number of feature maps ($L=4$ in our experiments).
%Following~\cite{zhao2024single}, this first block of the proposed framework learns individual joint representations before modeling their interdependencies, implementing a local-to-global hierarchy. This approach enhances scalability and modularity, as it avoids the heavy computation of globally modeling all elements simultaneously.

\tit{Deformable Surroundings Extraction} As shown in~\cite{luo2016understanding,zhao2023contextaware}, it is possible to encode fine-grained visual cues -- \ie, the joint locations -- and extract high-level semantics, \ie, the spatial configuration of the joints, via the high- and low-level features of a stacked network based on down-sampling operations~\cite{sun2019deep,yuan2021hrformer}.
Therefore, following~\cite{zhao2023contextaware}, we leverage a Deformable Context Extraction (DCE) module, based on the deformable attention mechanism~\cite{zhudeformable}. DCE extracts spatial contextual cues from the intermediate feature maps using the initial 2D pose joints as reference points.
% , suitably readapted to be referred to the cropped frame.
Linear projections of the 2D input poses are concatenated to the hierarchical features $\mathcal{H}$ as an additional channel. 
% FINO A QUI

\definecolor{myrowcolor}{HTML}{F5F5F5}
\begin{table*}[t!]
\centering
% \rowcolors{2}{gray!10}{white}
\begin{tabular}{@{}lccccc|c@{}}
\toprule
\textbf{Method}        & \textbf{Office}             & \textbf{Living Room}                 & \textbf{Kitchen}            & \textbf{Library}            & \textbf{Courtyard}          & \textbf{All}                \\ \midrule
\textit{Fixed bias}    & $88.0/76.0$          & $85.5/76.7$          & $86.0/82.4$          & $89.0/85.1$          & $89.7/88.7$          & $88.1/79.7$          \\
\textit{Frontal gaze}    & $22.6/21.9$          & $36.6/35.4$          & $\mathbf{17.9}/19.6$          & $27.1/25.8$          & $30.5/33.8$          & $28.8/28.8$     \\ \midrule
Dias \textit{et al.}~\cite{dias2020gaze}    & $-/27.2$             & $-/25.2$             & $-/19.8$             & $-/24.9$             & $-/36.1$             & $-/27.1$             \\
XGaze~\cite{zhang2020eth}         & $24.2/23.0$          & $42.0/40.9$          & $23.3/22.9$          & $24.6/22.3$          & $30.2/31.9$          & $29.2/28.4$          \\
Nonaka \textit{et al.}~\cite{nonaka2022dynamic}          & $20.0/18.1$          & $25.6/25.5$          & $21.5/18.6$          & $21.9/20.1$          & $28.4/30.5$          & $24.1/23.3$          \\ \midrule
Gaze360~\cite{kellnhofer2019gaze360}$^\dagger$       & $24.0/19.2$          & $41.1/31.3$          & $32.4/21.2$          & $27.5/20.7$          & $28.2/28.3$          & $30.4/24.5$          \\
Nonaka \textit{et al.}~\cite{nonaka2022dynamic}$^\dagger$ & $\mathbf{14.4/14.3}$ & $25.1/22.6$          & $20.4/19.6$          & $19.8/18.4$          & $25.4/\mathbf{26.9}$          & $21.7/20.9$          \\ \midrule
\textbf{Ours}$_{\text{(H=20, A=AVG)}}$          & $15.8/16.3$          & $\mathbf{19.3/20.6}$ & $18.2/\mathbf{19.5}$ & $\mathbf{17.6/16.9}$ & $\mathbf{25.3}/29.1$ & $\mathbf{19.5/20.5}$ \\ 
\textbf{Ours}$_{\text{(H=20, A=ORC$_{\text{G}}$)}}$     & $11.6/11.6$  & $13.2/13.8$ & $14.6/13.7$ & $14.2/12.9$ & $23.9/27.9$ & $15.9/16.3$ \\ \bottomrule
\end{tabular}
\caption{Experimental results on GAFA dataset expressed as MAE$_{\text{3D}}$$/$MAE$_{\text{2D}}$. $^\dagger$ indicates methods leveraging temporal information.}
\label{tab:gafa_results}
\end{table*}

The output of the DCE module $F_p' \in \mathbb{R}^{(L+1) \times J \times d}$ is an embedding containing near context (\ie, surroundings) and body pose features. We fixed $d=128$ in our experiments. A linear projection of the noisy 3D poses at the current timestep coming from the diffusion scheduler (see Sect.~\ref{subsec:diffusion}) is then concatenated to $F_p'$. Moreover, a positional encoding of the diffusion timestep is added in order to generate  $F_p \in \mathbb{R}^{(L+2) \times J \times d}$ and to make the model aware of the current diffusion step. 

\noindent \textbf{Pose-to-Context and Joint-to-Joint Modules.}
Drawing inspiration from multi-modality models~\cite{kim2021vilt,bao2022vlmo} that employ a transformer encoder, we use a similar architecture to learn a joint representation. 
$F_p$ is a multichannel descriptor, which contains two channels for the pose and $L$ channels for the context. The Pose-to-Context Attention Module performs a self-attention among the $L+2$ descriptors (tokens) of size $d$ for each joint. 
The Joint-to-Joint Attention Module considers $J$ tokens of size $d' = d \cdot (L+2)$ and computes self-attention among them. The Pose-to-Context module enriches the embeddings of each joint with contextual information, while the Joint-to-Joint module enables data sharing between the different joints.
$BS_p \in \mathbb{R}^{d' \times J}$ is the final output of the Body \& Surroundings module, with a descriptor of size $d'$ for each joint.

\subsection{Context with Objects}
\label{subsec:objectToPose}
The goal of this module is to extract information from the whole image, particularly focusing on elements that can affect or guide the person's gaze -- \ie, the objects.
To this aim, we use a DETR-like object detector, which is able to provide a descriptor of the objects in the image, with knowledge of both their location and their class. 
Let $F_{DETR} \in \mathbb{R}^{Q \times d'}$ be the last hidden states (removing the localization and classification heads and projecting to the common size $d'$) of the detector obtained with $Q$ input queries.
The Object-To-Context block performs a cross-attention between $F_{DETR}$ and $BS_p$. An average pooling operation is applied along the query dimension to merge all the important information related to the scene objects. 
% \davide{I think we should enrich this part making it more catchy and providing motivation, also connecting it to the clear improvements shown in the ablation study. This is one of our main contributions: using scene context to get 3D gaze from single RGB.} 
The obtained embedding $CO \in \mathbb{R}^{d'}$ is merged with $BS_p ^{Gaze}$ to generate the final Pose\&Gaze embedding $PG \in \mathbb{R}^{ d' \times J}$.

\subsection{Diffusion-based Multi-hypothesis Generation} \label{sec:diffusion}
\label{subsec:diffusion}
Estimating the 3D gaze and pose of people from RGB is inherently challenging. Major issues are the partial or complete occlusion of the eyes and the lack of depth information. Therefore, we propose the use of diffusion models to estimate the gaze direction, as their ability to generate multiple plausible hypotheses based on the person's pose and contextual information becomes highly valuable. By modeling various potential gaze directions, the diffusion process accommodates the inherent uncertainties and ambiguities. 
% For these reasons, the final predictions of \method are produced by a diffusion model. 

Representing pose and gaze direction using a single skeleton with an additional joint brings two advantages.
First, it enables the formalization of the global inference process as a denoising task, starting from a completely random pose sampled from a unique Gaussian distribution.
Second, it simplifies the optimization function: we adopted a single MSE loss between the predicted and the ground-truth joint coordinates, implicitly incorporating and standardizing the contributions of the pose and the gaze.

The iterative denoising procedure can be applied in parallel to $H$ initial hypotheses $\hat{y}_{N,h} \sim \mathcal{N}(0;1)$ in order to generate $H$ final predictions $\hat{y}_{0,h} \in \mathbb{R}^{J \times 3}$ after $N$ denoising iterations. 
For efficient inference, we employ the optimized DDIM~\cite{songdenoising} denoising scheduler.
A regression head is included in the diffusion module, and it is trained to perform the denoising task.

% Prior work on 3D human pose estimation using diffusion models~\cite{gong2023diffpose,shan2023diffusion,choi2023diffupose} has adapted successful 2D-to-3D lifting architectures as denoisers, drawing inspiration from the lifting literature. These lifting methods~\cite{zheng20213d,li2022mhformer} typically predict the 3D pose of a central frame using a sequence of 2D poses. However, this non-causal approach requires future 2D poses, making it unsuitable for real-time applications. Furthermore, it necessitates a tracking system for multi-person scenarios and introduces significant latency.

\noindent \textbf{Gaze and Pose aggregation.}
% One key advantage of a diffusion-based approach like \method~is its ability to integrate the simplicity of deterministic training with the flexibility of probabilistic inference.  
% As discussed in Section~\ref{subsec:diffusion}, 
\method generates $H$ hypotheses of the gaze and the pose, each one representing a plausible 3D solution. The distribution itself contains additional information about the real gaze (and posture). Therefore, a correct aggregation of the generated hypotheses allows for reducing the prediction error of a single hypothesis.

As the aggregation function $A$, we adopt the average operation (\textbf{AVG}) at joint level. Despite its simplicity, this aggregation has proven to be effective and accurate, as reported in experiments. 
For the sake of completeness, we also compute the ``Supervision from an Oracle''~\cite{sharma2019monocular} aggregation (\textbf{ORC}$_\text{G}$) that selects the closest hypothesis with respect to the ground truth annotation. This aggregation is useful to highlight the upper-bound performance of the proposed method, but it is limited in its applicability 
% in real-world scenarios 
when ground truth annotations are not available.
Additional aggregation functions based on oracle are investigated in Section~\ref{sec:ablation}.
We avoided using additional aggregation techniques that required ground-truth information or camera calibration parameters \cite{shan2023diffusion}, which are not always available or predictable in single-frame methods.
\section{Experimental Evaluation}

\subsection{Datasets}
As \method is based on the rich information extracted from the surroundings and global context, we focus on datasets containing 3D gaze annotations and full images, thus excluding those only containing crops around the faces, eyes or body of the subject~\cite{kellnhofer2019gaze360,fischer2018rt,zhang2020eth}. Additional details about datasets are reported in the Supplementary.

\noindent \textbf{GAFA}~\cite{nonaka2022dynamic} (Gaze from Afar) dataset is designed for 3D gaze estimation in surveillance scenarios, capturing freely moving people in natural settings. It includes more than $850$k video frames from 5 different daily environments. It features a wide range of head poses, including back views and high-pitch angles, reflecting realistic conditions. GAFA is annotated with 3D gaze directions and body orientations, using wearable cameras and AR marker-based positioning systems for ground truth.

\noindent \textbf{GFIE}~\cite{hu2023gfie} is a dataset introduced for 2D and 3D gaze-following tasks, created using a system that combines a laser rangefinder and an RGB-D camera to record and annotate gaze behaviors in natural indoor environments. The system guides the subject's gaze target using a laser spot, which is then detected in the RGB images to generate precise annotations, and then removed using image inpainting~\cite{ulyanov2018deep}. The 3D gaze target is reconstructed using the distance measured by the laser rangefinder and the camera's intrinsic parameters. The dataset includes about $71$k frames of $61$ subjects, performing a wide range of activities.

\noindent \textbf{Ego-Gaze}. We create this dataset starting from the multimodal Ego-Exo4D dataset~\cite{grauman2024ego}. Specifically, we select frames from the Ego-Pose subset in which the 3D annotation of the human pose is available. Then, we compute 3D gaze annotations from the data acquired with the Aria glasses. The dataset includes a wide range of skilled activities--- such as sports, music, dance---performed in natural settings. Because the Ego-Pose dataset is still used for competitions, the official test set has not been made available. Therefore, we use the official validation split as test set and we sample validation instances from the training set. We will release splits, enabling future comparisons.

\begin{figure*}[th]
    \centering
    \includegraphics[width=1\linewidth, page=2]{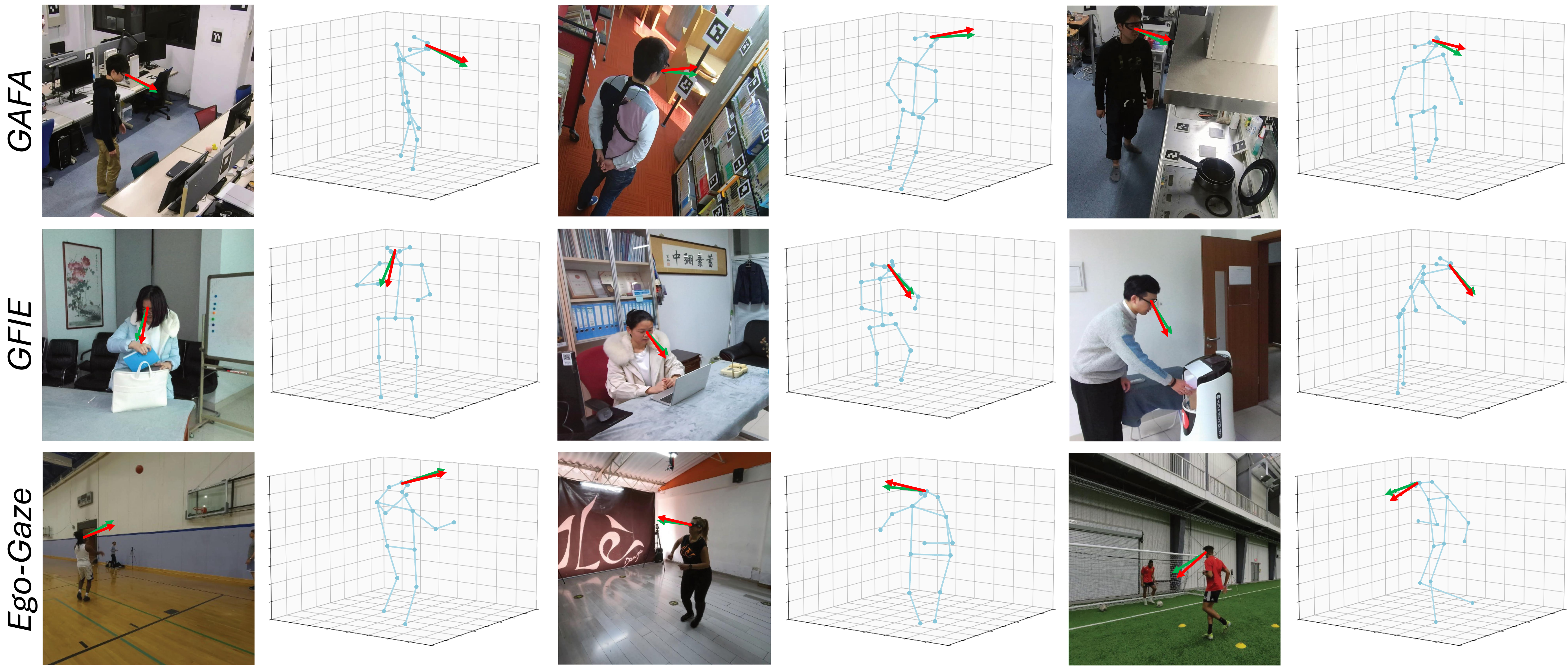}
    \caption{Qualitative Results on the three datasets. 
    % We present the input RGB image alongside the predicted 3D gaze and 3D pose. 
    \textcolor{darkgreen}{Green} arrow represents ground truth gaze direction,
    % annotation is shown in , 
    \textcolor{red}{red} arrow is the prediction
    % is displayed in .
    }
    \label{fig:qualitative}
\end{figure*}

% \begin{figure}[th]
%     \centering
%     \includegraphics[width=\linewidth, page=3]{images/qualitative.pdf}
%     \caption{Qualitative results, proposal 2}
%     \label{fig:qualitative}
% \end{figure}

\begin{table}[t!]
\centering
% \rowcolors{2}{gray!10}{white}
\begin{tabular}{@{}lcccc@{}}
\toprule
\textbf{Method}     & \textbf{RGB} & \textbf{Crops} & \textbf{Depth} & \textbf{MAE$_{\text{3D}}$} \\ \midrule
\textit{Random}     &  &  &  & $84.4$ \\ 
\textit{Center}     &  &  &  & $87.2$ \\ \midrule
GazeFollow~\cite{recasens2015they}                  & \ding{51} & \ding{51} & & $41.5$ \\
Lian \textit{et al.}~\cite{lian2018believe}         & \ding{51} & \ding{51} &  & $26.7$ \\
Rt-Gene~\cite{fischer2018rt}                        & \ding{51} & \ding{51} &  & $21.0$ \\
Hu \textit{et al.}~\cite{hu2023gfie}                & \ding{51} & \ding{51} & \ding{51} & $17.7$ \\
Toaiari \textit{et al.}~\cite{toaiari2024upper}     &  &  & \ding{51} & $15.9$ \\ \midrule
Chong \textit{et al.}~\cite{chong2020detecting}$^\dagger$      & \ding{51} & \ding{51} & & $20.8$ \\
Gaze360~\cite{kellnhofer2019gaze360}$^\dagger$      & \ding{51} &  & & $19.8$ \\ \midrule 
\textbf{Ours}$_{\text{(H=20, A=AVG)}}$              & \ding{51} & &  & $\mathbf{13.6}$ \\
\textbf{Ours}$_{\text{(H=20, A=ORC$_{g}$)}}$        & \ding{51} & &  & $9.9$ \\ \bottomrule
\end{tabular}
\caption{Quantitative results on GFIE dataset. For each method, the input data is reported: \textbf{RGB} for color images, \textbf{Crops} for head, face, or eye crops, and \textbf{Depth} for depth maps. $^\dagger$ indicates methods leveraging temporal information.}
\label{tab:gfie_results}
\end{table}

\begin{table}[th!]
\centering
\resizebox{\linewidth}{!}{
\begin{tabular}{@{}lcccc@{}}
\toprule
\textbf{Method}     & \textbf{Basket} & \textbf{Dance} & \textbf{Various} & \textbf{All} \\ \midrule
XGaze      & $21.6/20.6$ & $23.8/27.1$ & $21.6/20.5$ & $23.1/24.9$ \\
Gaze360    & $21.8/17.0$ & $21.0/21.8$ & $22.1/17.6$ & $21.3/20.4$ \\
\textbf{Ours}        & $\mathbf{15.4/14.7}$ & $\mathbf{18.6/18.9}$ & $\mathbf{15.3/12.7}$ & $\mathbf{17.5/17.4}$ \\ \bottomrule
\end{tabular}}
\caption{Quantitative results on the Ego-Gaze dataset. \method is tested with H=20, A=AVG.}
\label{tab:egoexo_results}
\end{table}

\subsection{Implementation Details and Training}
\label{subsec:implementation}
As backbones, we use different pre-trained models. 
For 2D human pose estimation, we use HRNet~\cite{sun2019deep}, capable of maintaining high-resolution representations throughout the whole architecture and achieving great accuracy.
As an object detector, we use RT-DETR~\cite{zhao2024detrs}, a recent end-to-end architecture with good accuracy and real-time performance for the object detection task from single RGB images.  
Both models are frozen during the training phase and are used with their original weights and parameters.

We train {\method} with a batch size of $64$ for $100$ epochs on all datasets.
We use Adam optimizer~\cite{kingma2014adam} with a starting learning rate of $6e^{-4}$ using a linear decay with factor $0.993$. No data augmentation is applied to input images. 

\begin{figure}[t!]
    \centering
    \begin{subfigure}[t]{0.49\linewidth}
        \centering
        \includegraphics[width=\linewidth]{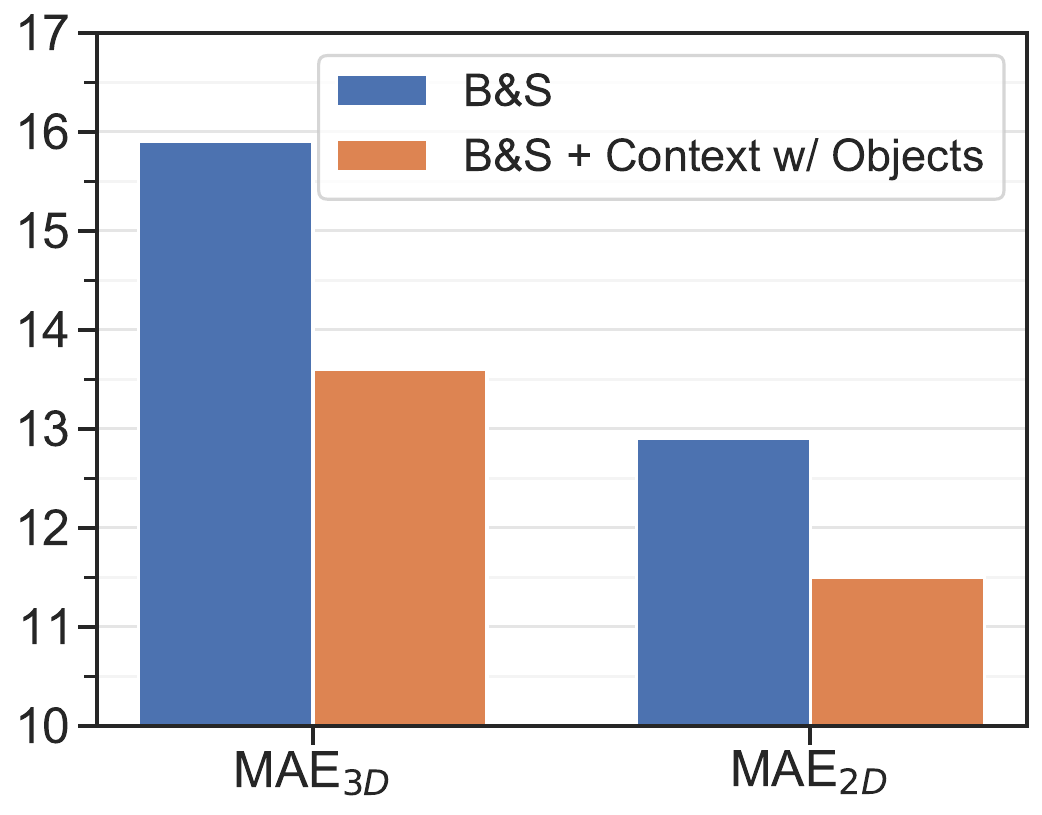}
        \caption{}
        \label{fig:abl_objects}
        \vspace*{2mm}
    \end{subfigure}
    \begin{subfigure}[t]{0.49\linewidth}
        \centering
        \includegraphics[width=\linewidth]{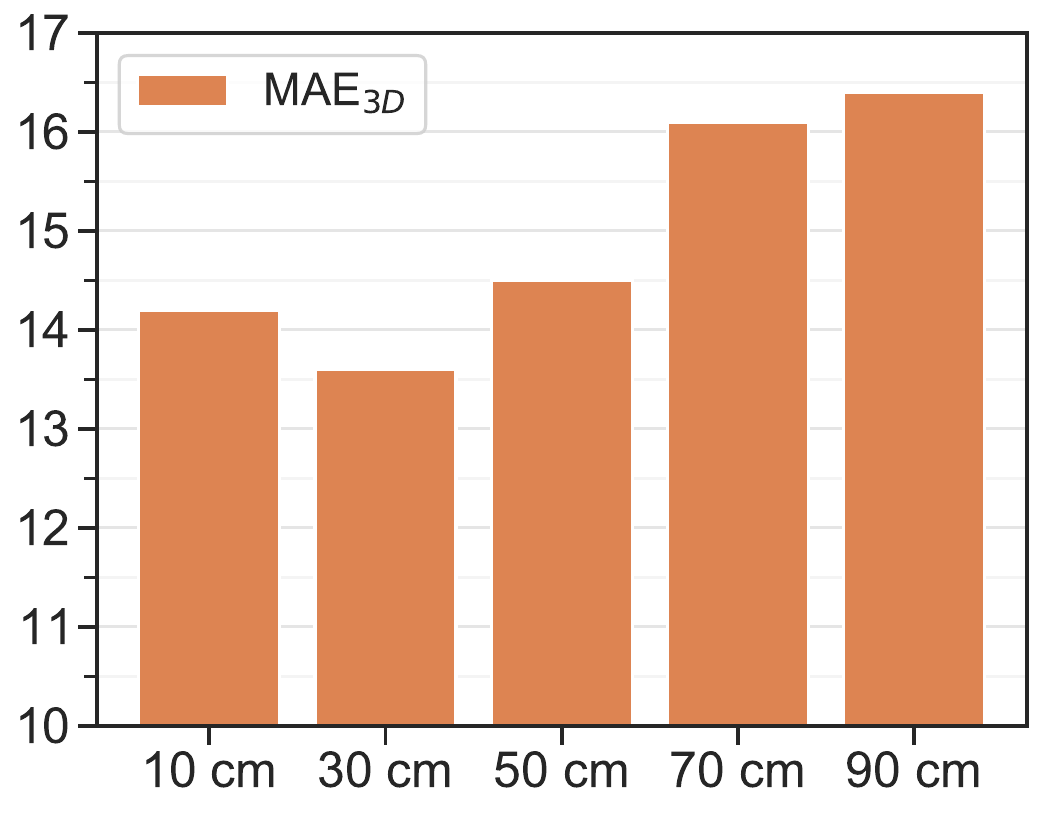}
        \caption{}
        \label{fig:abl_2}
    \end{subfigure}
    \caption{Ablation study on GFIE dataset: (a) the contribution of the  ``Context with objects'' module in addition to the ``Body\&Surroundings''. 
    % \davide{I think that for ablation (a) a table is better. You can see there is an improvement but it's not clear how much}. 
    (b) Performance of \method in terms of  MAE$_{\text{3D}}$ by varying the distance of the gaze joint from the head.}
    \label{fig:distances}
\end{figure}

\subsection{Baselines and Competitors}
We compare \method with several 3D gaze estimation baselines and competitors.
% and literature competitors developed for the 3D gaze estimation task. 

For the GAFA dataset~\cite{nonaka2022dynamic}, we compute two baselines. 
The first is \textit{fixed bias}, \ie the mean gaze direction is obtained from the training set, and the error metric is computed using this mean value over the test set~\cite{nonaka2022dynamic}. This baseline is intended to show the lower bound accuracy on this dataset.
The second baseline is \textit{frontal gaze}, where we compute the angular error assuming that the predicted gaze direction is always orthogonal to the line between the two eyes. This is a useful reference for understanding 
the precision that a method based solely on the pose of the head would achieve.
%whether a method predicts the mere head pose or not.
As competitors, we use a variety of state-of-the-art methods from the literature.
The approach proposed by Dias \textit{et al.}~\cite{dias2020gaze} estimates 2D gaze on the image plane using facial keypoints detected by OpenPose~\cite{cao2019openpose}. Gaze360~\cite{kellnhofer2019gaze360} takes a sequence of full-head images as input and provides the 3D gaze direction. XGaze~\cite{zhang2020eth} uses facial images as input and assumes high-resolution facial images. 

% All these methods are recent and represent the current state of the art on the GAFA dataset.

\begin{figure}[th!]
     \centering
     \begin{subfigure}[b]{0.5\linewidth}
         \centering
         \vstretch{1.3}{\includegraphics[width=\textwidth]{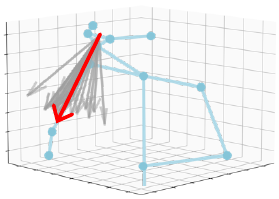}}
         \caption{}
         \label{fig:multiple_hyp}
     \end{subfigure}
     \medspace
     \begin{subfigure}[b]{0.45\linewidth}
         \centering
         \vstretch{0.8}{\includegraphics[width=\textwidth]{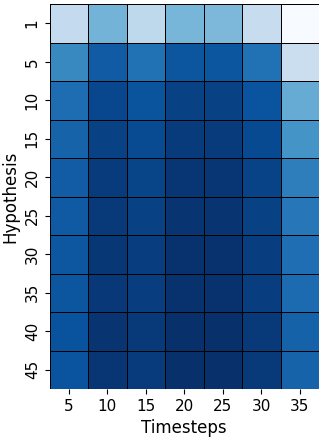}}
         \caption{}
         \label{fig:matrix_timesteps}
     \end{subfigure}
        \caption{ (a) \method, 
        % based on a diffusion architecture, 
        predicts multiple hypotheses on which aggregation functions are applied (see Sect.~\ref{sec:diffusion}) (b) Investigation on the number of different hypotheses vs number of timesteps. Darker color represents a lower MAE$_{3\text{D}}$ value.}
    \label{fig:tmp}
    %\vspace{-.1cm}
\end{figure}

\begin{table}[t!]
\centering
    \resizebox{\linewidth}{!}{
    \begin{tabular}{c c c c|c c}
    \toprule
    \multirow{2}{*}{\textbf{\#}} &
    \multirow{2}{*}{\textbf{Diffusion}} &
    \multirow{2}{*}{\textbf{Objects}} &
    \textbf{Near} &
    \multirow{2}{*}{\textbf{MAE$_{\text{3D}}$}} &
    \multirow{2}{*}{\textbf{MAE$_{\text{2D}}$}} \\ 
     &&&\textbf{Context} &&\\
     \midrule
    1 & {\color{red}\ding{55}} & {\color{green}\ding{51}} & {\color{green}\ding{51}} & 16.1 & 14.2 \\
    2 & {\color{green}\ding{51}} & {\color{red}\ding{55}} & {\color{green}\ding{51}} & 15.9 & 12.9 \\
    3 & {\color{green}\ding{51}} & {\color{green}\ding{51}} & {\color{red}\ding{55}} & 15.1 & 12.0 \\
    4 & {\color{green}\ding{51}} & {\color{green}\ding{51}} & {\color{green}\ding{51}} & 13.6 & 11.5 \\ 
    \bottomrule
    \end{tabular}
    }
    \caption{Ablation analysis of each component.}
    \label{tab:ablative}
\end{table}

\begin{table}[t!]
    \centering
    \begin{tabular}{ccc|cc}
        \toprule 
        \textbf{Aggregation} & \multicolumn{2}{c}{\textbf{GAFA~\cite{nonaka2022dynamic}}} & \multicolumn{2}{c}{\textbf{GFIE~\cite{hu2023gfie}}} \\
        \textbf{method} & MAE$_{\text{3D}}$ & MAE$_{\text{2D}}$ & MAE$_{\text{3D}}$ & MAE$_{\text{2D}}$ \\
        \midrule
        AVG & $19.5$ & $20.5$ & $13.6$ & $11.5$ \\
        ORC$_{\text{P}}$  & $19.7$ & $20.4$ & $13.4$ & $10.8$ \\
        ORC$_{\text{G}}$ & $15.9$ & $16.3$ & $9.9$ & $7.9$ \\
        ORC$_{\text{J}}$ & $12.6$ & $13.3$ & $8.7$ & $7.5$ \\
        \bottomrule
    \end{tabular}
    \caption{Impact of different aggregation methods on gaze.}
    \label{tab:gafa_gfie}
\end{table}

The GFIE dataset was originally proposed for gaze target detection, and thus provides additional information on the scene -- \ie depth maps. For this reason, we evaluate on this dataset with other baselines and competitors. 
The baseline methods include the \textit{random} approach, \ie the 2D and 3D gaze directions are randomly selected within the image and point cloud, respectively. In addition, the \textit{center} baseline localizes the gaze always at the center of the point cloud of the 3D space.
As competitors, we also use existing 2D gaze-following methods, \ie GazeFollow~\cite{recasens2015they}, Lian~\cite{lian2018believe}, and Chong~\cite{chong2020detecting}. To retrieve their 3D gaze angle, we first 
back-projected the 2D gaze target into the 3D space using the available registered depth maps. 
The results of Gaze360~\cite{kellnhofer2019gaze360} and Rt-Gene~\cite{fischer2018rt} are collected from~\cite{hu2022we}.
% the work by Hu \etal 
Finally, we report the results of the recent work by Toaiari \etal ~\cite{toaiari2024upper}, which utilizes upper-body skeleton data and the depth map of the scene to predict the 3D gaze.

\subsection{Comparison with state-of-the-art}
We report the result using the Mean Angular Error (MAE), the standard metric for the evaluation of gaze estimation methods. MAE is expressed in degrees, and it is calculated as the average of the angular difference between the predicted and ground-truth gaze directions over all the testing samples. In addition to the 3D errors (MAE$_{\text{3D}}$), we report the metric using the directions on the image plane (MAE$_{\text{2D}}$).

Table~\ref{tab:gafa_results} reports the performance of our \method and other approaches. On the GAFA dataset, as shown, \method achieves the best performance on both metrics, even outperforming methods that leverage additional temporal information, \ie~\cite{kellnhofer2019gaze360,nonaka2022dynamic}.
Both MAE$_{\text{3D}}$ and MAE$_{\text{2D}}$ achieved by our method are well below the \textit{frontal gaze} baseline, indicating that the idea to model the gaze as additional joint is effective in estimating the 3D gaze direction and that the output of our method is not the mere head pose.

Table~\ref{tab:gfie_results} reports similar results on the GFIE dataset.
Also in this case, our method largely outperforms the competitors. 
In particular, \method outperforms even methods that are based on fine details, such as the face or body crops, or additional input data as depth maps, whose contribution is significant in 3D estimation tasks.

\begin{table*}[th!]
    \centering
    \resizebox{\linewidth}{!}{
    \begin{tabular}{l | c c c c c c c c c c c | c c }
        \toprule
        \textbf{Methods} & Mart. & Zhao & Sun & Yang & Hoss. & Liu & Xu & Zhao & Zhao  & Diffu. & Diffp. & \textbf{Ours} & \textbf{Ours}\\
          & \cite{martinez2017simple} & \cite{zhaoCVPR19semantic} & \cite{sun2017compositional} & \cite{yang20183d} & \cite{hossain2018exploiting} & \cite{liu2020comprehensive} & \cite{xu2021graph} & \cite{zhao2022graformer} & \cite{zhao2023contextaware}  & \cite{choi2023diffupose} & \cite{gong2023diffpose} & {\small (H=20, A=AVG)}  &{ \small (H=20, A=ORC)} \\
        \midrule
        \textbf{MPJPE} $\downarrow$ &62.9 &60.8 &59.1 &58.6 &58.3 &52.4 &51.9 &51.8 &\textbf{43.4} &\underline{49.4} &49.7 & 49.7 & 41.1 \\
        \bottomrule
    \end{tabular}
    }
    \caption{Results on Human3.6M dataset for the 3D Human Pose Estimation task. The best result is in \textbf{bold}, the second one is \underline{underlined}.}
    \label{tab:hpe_h36}
\end{table*}

\begin{table*}[th!]
\centering
\begin{tabular}{c | c c c c c c c | c c}
% {r c|c x{1.5cm} x{1.5cm} x{1.5cm} x{1.5cm} x{1.5cm} x{1.5cm} x{1.5cm} c|c x{2cm} x{2cm}}
\toprule
\textbf{Methods} & Pavvlo &  Zheng & Wang & Li & Zheng & Zhang  & Zhao 
& \textbf{Ours}  & \textbf{Ours} \\
 & \cite{pavllo20193d} &  \cite{zheng20213d} & \cite{wang2020motion} & \cite{li2022mhformer} & \cite{zheng20213d} & \cite{zhang2022mixste} & \cite{zhao2023contextaware} 
 & {\small (H=20, A=AVG)}  &{ \small (H=20, A=ORC)} \\
\midrule
\textbf{MPJPE $\downarrow$} & 84.0 & 77.1 & 68.1 & 58.0 & 57.7 & 54.9 & \textbf{44.7} & \underline{46.6}  & 33.8 \\
\bottomrule
\end{tabular}
\caption{Results on MPI-INF-3DHP dataset for the 3D Human Pose Estimation task. The best result is in \textbf{bold}, the second one is \underline{underlined}.}
 \vspace{-0.5em}
\label{tab:hpe_mpi}
\end{table*}
In both datasets, we also report the results obtained using the ORC$_\text{G}$ aggregation function.
As expected, these are the best results. However, the ground truth is not normally available in the inference phase, and it is not completely correct to use it to select the best hypothesis. These results show that the diffusion process can generate hypotheses close to the ground truth, suggesting future work on more sophisticated aggregation strategies.

The newly introduced Ego-Gaze dataset imposes re-training the Gaze360 and XGaze methods. Unfortunately, it was not possible to implement more recent techniques, such as the ones developed in~\cite{nonaka2022dynamic,hu2023gfie}, due to a lack of depth maps and body or head orientation, respectively. 
Results are reported in Table~\ref{tab:egoexo_results}, organized in three main scenes, \ie, basket, dance and various. The latter includes the less represented classes, such as cooking, soccer and bike repair. 
As shown, \method achieves the best results in all the scenes.
These results demonstrate the robustness of the proposed approach on a challenging dataset with complex scenes.

\subsection{Qualitative results}
Some qualitative results are reported in Figure~\ref{fig:qualitative}, where the input image and the predicted 3D gaze and pose are shown. The ground truth gaze direction vector is drawn as a green arrow, while the predicted vector is drawn as a red arrow. These results confirm the ability of \method to predict gaze direction in wide-angle ranges, also when the face is not visible or partially occluded.
% (\eg, when the subject is turned away).
Additional qualitative results are reported in the Supplementary material.

\subsection{Ablation Studies} \label{sec:ablation}
Ablation studies are mainly computed on the GFIE dataset, using \method in the configuration described in Section \ref{subsec:implementation}.
\\
\noindent \textbf{Module contributions} We investigate the contribution of each module (see Table \ref{tab:ablative}). In experiment \#1, we use the transformer-based model without the diffusion process, training the network to predict directly pose and gaze. In \#2, we remove the module ''Context with Objects'' for context analysis. In \#3, we remove the part of the method responsible for extracting and processing the surroundings of the person. 
% As shown, 
Each module is a key part of the method.
% and contributes positively on the final result.
\\
\noindent \textbf{Context with Objects module} 
To highlight the performance improvement provided by the proposed ``context with objects'' module, we tested the results of \method directly using $BS_p$ in input to the diffusion step (see Fig.~\ref{fig:general}). The results are reported in Figure~\ref{fig:abl_objects}.
% both in terms of MAE$_{\text{3D}}$ (left) and MAE$_{\text{2D}}$ (right).
Adding the object embeddings clearly improves the model's ability to solve the 3D gaze estimation task. 
\\
\noindent \textbf{Distances of the Gaze Keypoint} The gaze joint is an auxiliary point used to solve the task, but it is not physically present. Its distance from the eyes was chosen to be close enough to the body to be modeled as a joint and, at the same time, far enough to reduce the dependence on noise in the final conversion into angles. The measurement used, equal to $30$ centimeters, is supported by an experimental analysis.
MAE$_{\text{3D}}$ errors vs
% obtained using different 
distances are plotted in Figure~\ref{fig:abl_2}. 
\\
\noindent \textbf{Number of hypotheses and timesteps} A key advantage of diffusion models is their ability to generate multiple hypotheses. In Figure~\ref{fig:multiple_hyp}, a real multiple-hypothesis prediction is depicted. 
% In this ablation study, 
Then, we analyze how the number of hypotheses $H$ and 
the number of denoising iterations $N$ affect the final MAE$_{\text{3D}}$. Figure~\ref{fig:matrix_timesteps} shows the matrix from which we selected the final values of $H$ and $N$, where darker colors denote lower errors.
Based on this analysis, we selected $H=20$ and $N=20$ for our 
% experimental 
evaluation, as a favorable trade-off between accuracy and computational load. 
% The  experiment has been done on the test set of GFIE dataset by averaging the predictions across the generated hypotheses.
\\
\noindent \textbf{Multiple Hypothesis Aggregation Strategies} Having multiple generated hypotheses allows us to explore various aggregation strategies.
In Table~\ref{tab:gafa_gfie}, we compare the different aggregations described in Section~\ref{sec:diffusion}. 
We also investigate different oracle selections~\cite{sharma2019monocular} 
% (ORC metric, hereinafter) 
obtained in three different ways. 
% As mentioned, 
ORC$_{\text{G}}$ chooses the hypothesis with the lowest error, specifically at the gaze joint.
ORC$_{\text{P}}$ selects the hypothesis with the lowest Mean Per-Joint Position Error (MPJPE) relative to the ground truth; however, our results indicate that minimizing MPJPE at the pose level does not necessarily produce the most accurate gaze estimation. 
Finally, ORC$_{\text{J}}$ employs a per-joint selection strategy in which, for each joint, the coordinates with the lowest error are independently selected, resulting in a more accurate estimation of gaze direction.
Since ORC$_{p}$, ORC$_{g}$, and ORC$_{j}$ rely on ground truth data, they are not applicable in real-world scenarios. Therefore, we consider \textbf{AVG} as the most appropriate baseline for fair comparison.

\subsection{Additional Evaluation} 
As previously mentioned, \method predicts not only the 3D gaze, but also the 3D body pose: therefore, we also analyze performances on this task pose. Therefore, in this section, we analyze the performance of this task. 
\\
\noindent \textbf{Dataset} The Human3.6M dataset~\cite{ionescu2013human3} is a well-known dataset of $3.6$ million images with 3D human pose annotations. 
It contains $17$-joint skeleton annotations for $11$ subjects performing $15$ activities, captured by $4$ cameras in an indoor environment. 
For evaluation, we follow the standard protocol of training on subjects S1, S5, S6, S7, and S8, and testing on subjects S9 and S11.
The MPI-INF-3DHP dataset consists of over $1.3$ million frames captured from $14$ cameras and it is widely used for training and evaluating 3D human pose estimation models. It contains 8 actors performing activities such as walking, sitting, and sports. The frames are annotated using a skeleton model with $17$ joints. 
\\
\noindent \textbf{3D Pose Evaluation} 
For the training, we use a batch of $128$ for $50$ epochs. Other training settings are the same used for the gaze evaluation. Performance is evaluated using the Mean Per Joint Position Error (MPJPE)~\cite{joo2015panoptic}, which calculates the average Euclidean distance (in millimeters) between predicted and ground truth 3D joint coordinates. 
In Tables~\ref{tab:hpe_h36} and~\ref{tab:hpe_mpi}, we report the comparison for the 3D pose estimation task between our model and literature competitors, on Human3.6M and MPI-INF datasets, respectively. Among the others, Diffupose~\cite{choi2023diffupose}, Diffpose~\cite{gong2023diffpose} are the most similar methods since they are based on a diffusion architecture. As shown, the results obtained are better than a large portion of the literature, and comparable with the most recent one.
These experimental results suggest that, although our method was not specifically developed for the HPE task, it still achieves competitive results with a good level of accuracy.

\section{Conclusion}
We introduced \method, a method for 3D gaze and pose estimation from single RGB images. By modeling 3D gaze through a diffusion process, \method integrates 2D pose, surrounding context, and global scene cues. The use of a diffusion model addresses the inherent ambiguity of 3D gaze estimation, generating multiple plausible hypotheses. Results demonstrate the efficacy of \method, highlighting its potential for accurate 3D gaze and pose estimation.

{
    \small
    \bibliographystyle{ieeenat_fullname}
    \bibliography{main}

@String(CVPR= {IEEE Conf. Comput. Vis. Pattern Recog.})

@String(ICCV= {Int. Conf. Comput. Vis.})

@String(ECCV= {Eur. Conf. Comput. Vis.})

@String(NIPS= {Adv. Neural Inform. Process. Syst.})

@String(ICPR = {Int. Conf. Pattern Recog.})

@String(ICME = {Int. Conf. Multimedia and Expo})

@String(ICLR = {Int. Conf. Learn. Represent.})

@String(CVPR  = {CVPR})

@String(ICCV  = {ICCV})

@String(ECCV  = {ECCV})

@String(NIPS  = {NeurIPS})

@String(ICPR  = {ICPR})

@String(ICME  =	{ICME})

@String(ICLR  = {ICLR})

@inproceedings{zhudeformable,
  title={Deformable DETR: Deformable Transformers for End-to-End Object Detection},
  author={Zhu, Xizhou and Su, Weijie and Lu, Lewei and Li, Bin and Wang, Xiaogang and Dai, Jifeng},
  booktitle=ICLR,
  year = {2020}
}

@article{luo2016understanding,
  title={Understanding the effective receptive field in deep convolutional neural networks},
  author={Luo, Wenjie and Li, Yujia and Urtasun, Raquel and Zemel, Richard},
  journal={Advances in neural information processing systems},
  volume={29},
  year={2016}
}

@inproceedings{sun2019deep,
  title={Deep high-resolution representation learning for human pose estimation},
  author={Sun, Ke and Xiao, Bin and Liu, Dong and Wang, Jingdong},
  booktitle=CVPR,
  year={2019}
}

@article{yuan2021hrformer,
  title={Hrformer: High-resolution vision transformer for dense predict},
  author={Yuan, Yuhui and Fu, Rao and Huang, Lang and Lin, Weihong and Zhang, Chao and Chen, Xilin and Wang, Jingdong},
  journal={Advances in neural information processing systems},
  volume={34},
  year={2021}
}

@inproceedings{kim2021vilt,
  title={Vilt: Vision-and-language transformer without convolution or region supervision},
  author={Kim, Wonjae and Son, Bokyung and Kim, Ildoo},
  booktitle={International conference on machine learning},
  year={2021},
  organization={PMLR}
}

@article{bao2022vlmo,
  title={Vlmo: Unified vision-language pre-training with mixture-of-modality-experts},
  author={Bao, Hangbo and Wang, Wenhui and Dong, Li and Liu, Qiang and Mohammed, Owais Khan and Aggarwal, Kriti and Som, Subhojit and Piao, Songhao and Wei, Furu},
  journal={Advances in Neural Information Processing Systems},
  volume={35},
  year={2022}
}

@inproceedings{zheng20213d,
  title={3d human pose estimation with spatial and temporal transformers},
  author={Zheng, Ce and Zhu, Sijie and Mendieta, Matias and Yang, Taojiannan and Chen, Chen and Ding, Zhengming},
  booktitle=ICCV,
  year={2021}
}

@inproceedings{songdenoising,
  title={Denoising Diffusion Implicit Models},
  author={Song, Jiaming and Meng, Chenlin and Ermon, Stefano},
  booktitle={International Conference on Learning Representations}
}

@inproceedings{d2021refinet,
  title={Refinet: 3d human pose refinement with depth maps},
  author={D'Eusanio, Andrea and Pini, Stefano and Borghi, Guido and Vezzani, Roberto and Cucchiara, Rita},
  booktitle={2020 25th International Conference on Pattern Recognition (ICPR)},
  pages={2320--2327},
  year={2021},
  organization={IEEE}
}

@inproceedings{li2022mhformer,
    title        = {Mhformer: Multi-hypothesis transformer for 3d human pose estimation},
    author       = {Li, Wenhao and Liu, Hong and Tang, Hao and Wang, Pichao and Van Gool, Luc},
    year         = 2022,
    booktitle    = CVPR,
    pages        = {13147--13156}
}

@inproceedings{gong2023diffpose,
    title        = {Diffpose: Toward more reliable 3d pose estimation},
    author       = {Gong, Jia and Foo, Lin Geng and Fan, Zhipeng and Ke, Qiuhong and Rahmani, Hossein and Liu, Jun},
    year         = 2023,
    booktitle    = CVPR,
}

@article{d2023depth,
  title={Depth-based 3D human pose refinement: Evaluating the refinet framework},
  author={D’Eusanio, Andrea and Simoni, Alessandro and Pini, Stefano and Borghi, Guido and Vezzani, Roberto and Cucchiara, Rita},
  journal={Pattern Recognition Letters},
  volume={171},
  pages={185--191},
  year={2023},
  publisher={Elsevier}
}

@inproceedings{shan2023diffusion,
  title={Diffusion-Based 3D Human Pose Estimation with Multi-Hypothesis Aggregation},
  author={Shan, Wenkang and Liu, Zhenhua and Zhang, Xinfeng and Wang, Zhao and Han, Kai and Wang, Shanshe and Ma, Siwei and Gao, Wen},
  booktitle=ICCV,
  year={2023}
}

@inproceedings{choi2023diffupose,
  author={Choi, Jeongjun and Shim, Dongseok and Kim, H. Jin},
  title={DiffuPose: Monocular 3D Human Pose Estimation via Denoising Diffusion Probabilistic Model}, 
  booktitle=IROS, 
  year={2023},
  pages={3773-3780}
}

@article{ionescu2013human3,
  title={Human3. 6m: Large scale datasets and predictive methods for 3d human sensing in natural environments},
  author={Ionescu, Catalin and Papava, Dragos and Olaru, Vlad and Sminchisescu, Cristian},
  journal={IEEE transactions on pattern analysis and machine intelligence},
  volume={36},
  number={7},
  year={2013},
  publisher={IEEE}
}

@inproceedings{joo2015panoptic,
  title={Panoptic studio: A massively multiview system for social motion capture},
  author={Joo, Hanbyul and Liu, Hao and Tan, Lei and Gui, Lin and Nabbe, Bart and Matthews, Iain and Kanade, Takeo and Nobuhara, Shohei and Sheikh, Yaser},
  booktitle=CVPR,
  year={2015}
}

@inproceedings{hu2023gfie,
  title={Gfie: A dataset and baseline for gaze-following from 2d to 3d in indoor environments},
  author={Hu, Zhengxi and Yang, Yuxue and Zhai, Xiaolin and Yang, Dingye and Zhou, Bohan and Liu, Jingtai},
  booktitle=CVPR,
  year={2023}
}

@inproceedings{ulyanov2018deep,
  title={Deep image prior},
  author={Ulyanov, Dmitry and Vedaldi, Andrea and Lempitsky, Victor},
  booktitle=CVPR,
  year={2018}
}

@inproceedings{nonaka2022dynamic,
  title={Dynamic 3d gaze from afar: Deep gaze estimation from temporal eye-head-body coordination},
  author={Nonaka, Soma and Nobuhara, Shohei and Nishino, Ko},
  booktitle=CVPR,
  year={2022}
}

@inproceedings{grauman2024ego,
  title={Ego-exo4d: Understanding skilled human activity from first-and third-person perspectives},
  author={Grauman, Kristen and Westbury, Andrew and Torresani, Lorenzo and Kitani, Kris and Malik, Jitendra and Afouras, Triantafyllos and Ashutosh, Kumar and Baiyya, Vijay and Bansal, Siddhant and Boote, Bikram and others},
  booktitle=CVPR,
  year={2024}
}

@inproceedings{sharma2019monocular,
    title        = {Monocular 3d human pose estimation by generation and ordinal ranking},
    author       = {Sharma, Saurabh and Varigonda, Pavan Teja and Bindal, Prashast and Sharma, Abhishek and Jain, Arjun},
    year         = 2019,
    booktitle    = ICCV,
}

@inproceedings{
    zhao2023contextaware,
    title={A Single 2D Pose with Context is Worth Hundreds for 3D Human Pose Estimation},
    author={Zhao, Qitao and Zheng, Ce and Liu, Mengyuan and Chen, Chen},
    booktitle=NIPS,
    year={2023},
}

@article{kingma2014adam,
    title        = {Adam: A method for stochastic optimization},
    author       = {Kingma, Diederik P and Ba, Jimmy},
    year         = 2014,
    journal      = {arXiv preprint arXiv:1412.6980}
}

@article{lu2016estimating,
  title={Estimating 3D gaze directions using unlabeled eye images via synthetic iris appearance fitting},
  author={Lu, Feng and Gao, Yue and Chen, Xiaowu},
  journal={IEEE Transactions on Multimedia},
  volume={18},
  number={9},
  year={2016},
  publisher={IEEE}
}

@inproceedings{nakazawa2012point,
  title={Point of gaze estimation through corneal surface reflection in an active illumination environment},
  author={Nakazawa, Atsushi and Nitschke, Christian},
  booktitle=ECCV,
  year={2012},
  organization={Springer}
}

@article{valenti2011combining,
  title={Combining head pose and eye location information for gaze estimation},
  author={Valenti, Roberto and Sebe, Nicu and Gevers, Theo},
  journal={IEEE Transactions on Image Processing},
  volume={21},
  number={2},
  year={2011},
  publisher={IEEE}
}

@inproceedings{hennessey2006single,
  title={A single camera eye-gaze tracking system with free head motion},
  author={Hennessey, Craig and Noureddin, Borna and Lawrence, Peter},
  booktitle={Proceedings of the 2006 symposium on Eye tracking research \& applications},
  year={2006}
}

@article{lee20123d,
  title={3D gaze tracking method using Purkinje images on eye optical model and pupil},
  author={Lee, Ji Woo and Cho, Chul Woo and Shin, Kwang Yong and Lee, Eui Chul and Park, Kang Ryoung},
  journal={Optics and Lasers in Engineering},
  volume={50},
  number={5},
  year={2012},
  publisher={Elsevier}
}

@inproceedings{zhu2005eye,
  title={Eye gaze tracking under natural head movements},
  author={Zhu, Zhiwei and Ji, Qiang},
  booktitle=CVPR,
  volume={1},
  year={2005},
  organization={IEEE}
}

@inproceedings{fischer2018rt,
  title={Rt-gene: Real-time eye gaze estimation in natural environments},
  author={Fischer, Tobias and Chang, Hyung Jin and Demiris, Yiannis},
  booktitle=ECCV,
  year={2018}
}

@inproceedings{zhang2020eth,
  title={Eth-xgaze: A large scale dataset for gaze estimation under extreme head pose and gaze variation},
  author={Zhang, Xucong and Park, Seonwook and Beeler, Thabo and Bradley, Derek and Tang, Siyu and Hilliges, Otmar},
  booktitle=ECCV,
  year={2020},
  organization={Springer}
}

@article{zhang2017mpiigaze,
  title={Mpiigaze: Real-world dataset and deep appearance-based gaze estimation},
  author={Zhang, Xucong and Sugano, Yusuke and Fritz, Mario and Bulling, Andreas},
  journal={IEEE transactions on pattern analysis and machine intelligence},
  volume={41},
  number={1},
  year={2017},
  publisher={IEEE}
}

@article{toaiari2024upper,
  title={Upper-Body Pose-based Gaze Estimation for Privacy-Preserving 3D Gaze Target Detection},
  author={Toaiari, Andrea and Murino, Vittorio and Cristani, Marco and Beyan, Cigdem},
  journal={arXiv preprint arXiv:2409.17886},
  year={2024}
}

@inproceedings{zhou2019learning,
  title={Learning a 3D gaze estimator with improved itracker combined with bidirectional LSTM},
  author={Zhou, Xiaolong and Lin, Jianing and Jiang, Jiaqi and Chen, Shengyong},
  booktitle=ICME,
  year={2019},
  organization={IEEE}
}

@article{sarbolandi2015kinect,
  title={Kinect range sensing: Structured-light versus Time-of-Flight Kinect},
  author={Sarbolandi, Hamed and Lefloch, Damien and Kolb, Andreas},
  journal={Computer vision and image understanding},
  volume={139},
  year={2015},
  publisher={Elsevier}
}

@article{cheng2024appearance,
  title={Appearance-based gaze estimation with deep learning: A review and benchmark},
  author={Cheng, Yihua and Wang, Haofei and Bao, Yiwei and Lu, Feng},
  journal={IEEE Transactions on Pattern Analysis and Machine Intelligence},
  year={2024},
  publisher={IEEE}
}

@article{palmero2018recurrent,
  title={Recurrent cnn for 3d gaze estimation using appearance and shape cues},
  author={Palmero, Cristina and Selva, Javier and Bagheri, Mohammad Ali and Escalera, Sergio},
  journal={arXiv preprint arXiv:1805.03064},
  year={2018}
}

@article{sitzmann2018saliency,
  title={Saliency in VR: How do people explore virtual environments?},
  author={Sitzmann, Vincent and Serrano, Ana and Pavel, Amy and Agrawala, Maneesh and Gutierrez, Diego and Masia, Belen and Wetzstein, Gordon},
  journal={IEEE transactions on visualization and computer graphics},
  volume={24},
  number={4},
  year={2018},
  publisher={IEEE}
}

@article{eckstein2017beyond,
  title={Beyond eye gaze: What else can eyetracking reveal about cognition and cognitive development?},
  author={Eckstein, Maria K and Guerra-Carrillo, Bel{\'e}n and Singley, Alison T Miller and Bunge, Silvia A},
  journal={Developmental cognitive neuroscience},
  volume={25},
  year={2017},
  publisher={Elsevier}
}

@inproceedings{raptis2017using,
  title={Using eye gaze data and visual activities to infer human cognitive styles: method and feasibility studies},
  author={Raptis, George E and Katsini, Christina and Belk, Marios and Fidas, Christos and Samaras, George and Avouris, Nikolaos},
  booktitle={proceedings of the 25th conference on user modeling, Adaptation and Personalization},
  year={2017}
}

@article{sharma2013eye,
  title={Eye gaze techniques for human computer interaction: A research survey},
  author={Sharma, Anjana and Abrol, Pawanesh},
  journal={International Journal of Computer Applications},
  volume={71},
  number={9},
  year={2013},
  publisher={Citeseer}
}

@article{kerr2019eye,
  title={Eye-tracking research in eating disorders: A systematic review},
  author={Kerr-Gaffney, Jess and Harrison, Amy and Tchanturia, Kate},
  journal={International Journal of Eating Disorders},
  volume={52},
  number={1},
  year={2019},
  publisher={Wiley Online Library}
}

@ARTICLE{Pose2Gaze,
  author={Hu, Zhiming and Xu, Jiahui and Schmitt, Syn and Bulling, Andreas},
  journal={IEEE Transactions on Visualization and Computer Graphics}, 
  title={Pose2Gaze: Eye-Body Coordination During Daily Activities for Gaze Prediction From Full-Body Poses}, 
  year={2024},
  volume={},
  number={},
  doi={10.1109/TVCG.2024.3412190}}

@article{vural2009eye,
  title={Eye-gaze based real-time surveillance video synopsis},
  author={Vural, Ulas and Akgul, Yusuf Sinan},
  journal={Pattern Recognition Letters},
  volume={30},
  number={12},
  year={2009},
  publisher={Elsevier}
}

@inproceedings{pal2020looking,
  title={"Looking at the right stuff"-Guided semantic-gaze for autonomous driving},
  author={Pal, Anwesan and Mondal, Sayan and Christensen, Henrik I},
  booktitle=CVPR,
  year={2020}
}

@inproceedings{palinko2015eye,
  title={Eye gaze tracking for a humanoid robot},
  author={Palinko, Oskar and Rea, Francesco and Sandini, Giulio and Sciutti, Alessandra},
  booktitle={2015 IEEE-RAS 15th International Conference on Humanoid Robots (Humanoids)},
  year={2015},
  organization={IEEE}
}

@inproceedings{tonini2023object,
  title={Object-aware gaze target detection},
  author={Tonini, Francesco and Dall'Asen, Nicola and Beyan, Cigdem and Ricci, Elisa},
  booktitle=ICCV,
  year={2023}
}

@inproceedings{cheng2018appearance,
  title={Appearance-based gaze estimation via evaluation-guided asymmetric regression},
  author={Cheng, Yihua and Lu, Feng and Zhang, Xucong},
  booktitle=ECCV,
  year={2018}
}

@inproceedings{funes2014eyediap,
  title={Eyediap: A database for the development and evaluation of gaze estimation algorithms from rgb and rgb-d cameras},
  author={Funes Mora, Kenneth Alberto and Monay, Florent and Odobez, Jean-Marc},
  booktitle={Proceedings of the symposium on eye tracking research and applications},
  year={2014}
}

@inproceedings{guo2020domain,
  title={Domain adaptation gaze estimation by embedding with prediction consistency},
  author={Guo, Zidong and Yuan, Zejian and Zhang, Chong and Chi, Wanchao and Ling, Yonggen and Zhang, Shenghao},
  booktitle={Proceedings of the Asian Conference on Computer Vision},
  year={2020}
}

@inproceedings{kellnhofer2019gaze360,
  title={Gaze360: Physically unconstrained gaze estimation in the wild},
  author={Kellnhofer, Petr and Recasens, Adria and Stent, Simon and Matusik, Wojciech and Torralba, Antonio},
  booktitle=ICCV,
  year={2019}
}

@inproceedings{guan2020enhanced,
  title={Enhanced gaze following via object detection and human pose estimation},
  author={Guan, Jian and Yin, Liming and Sun, Jianguo and Qi, Shuhan and Wang, Xuan and Liao, Qing},
  booktitle={MultiMedia Modeling: 26th International Conference, MMM 2020, Daejeon, South Korea, January 5--8, 2020, Proceedings, Part II 26},
  year={2020},
  organization={Springer}
}

@article{hu2022we,
  title={We know where they are looking at from the rgb-d camera: Gaze following in 3d},
  author={Hu, Zhengxi and Yang, Dingye and Cheng, Shilei and Zhou, Lei and Wu, Shichao and Liu, Jingtai},
  journal={IEEE Transactions on Instrumentation and Measurement},
  volume={71},
  year={2022},
  publisher={IEEE}
}

@article{guan2023end,
  title={End-to-end video gaze estimation via capturing head-face-eye spatial-temporal interaction context},
  author={Guan, Yiran and Chen, Zhuoguang and Zeng, Wenzheng and Cao, Zhiguo and Xiao, Yang},
  journal={IEEE Signal Processing Letters},
  volume={30},
  year={2023},
  publisher={IEEE}
}

@inproceedings{jindal2024spatio,
  title={Spatio-Temporal Attention and Gaussian Processes for Personalized Video Gaze Estimation},
  author={Jindal, Swati and Yadav, Mohit and Manduchi, Roberto},
  booktitle=CVPR,
  year={2024}
}

@inproceedings{cao2017realtime,
  title={Realtime multi-person 2d pose estimation using part affinity fields},
  author={Cao, Zhe and Simon, Tomas and Wei, Shih-En and Sheikh, Yaser},
  booktitle=CVPR,
  year={2017}
}

@inproceedings{chen2018cascaded,
  title={Cascaded pyramid network for multi-person pose estimation},
  author={Chen, Yilun and Wang, Zhicheng and Peng, Yuxiang and Zhang, Zhiqiang and Yu, Gang and Sun, Jian},
  booktitle=CVPR,
  year={2018}
}

@inproceedings{simo2012single,
  title={Single image 3D human pose estimation from noisy observations},
  author={Simo-Serra, Edgar and Ramisa, Arnau and Alenya, Guillem and Torras, Carme and Moreno-Noguer, Francesc},
  booktitle=CVPR,
  year={2012},
  organization={IEEE}
}

@inproceedings{oikarinen2021graphmdn,
  title={Graphmdn: Leveraging graph structure and deep learning to solve inverse problems},
  author={Oikarinen, Tuomas and Hannah, Daniel and Kazerounian, Sohrob},
  booktitle={2021 International Joint Conference on Neural Networks (IJCNN)},
  year={2021},
  organization={IEEE}
}

@article{ho2020denoising,
  title={Denoising diffusion probabilistic models},
  author={Ho, Jonathan and Jain, Ajay and Abbeel, Pieter},
  journal={Advances in neural information processing systems},
  volume={33},
  year={2020}
}

@inproceedings{zhao2024detrs,
  title={Detrs beat yolos on real-time object detection},
  author={Zhao, Yian and Lv, Wenyu and Xu, Shangliang and Wei, Jinman and Wang, Guanzhong and Dang, Qingqing and Liu, Yi and Chen, Jie},
  booktitle=CVPR,
  year={2024}
}

@article{recasens2015they,
  title={Where are they looking?},
  author={Recasens, Adria and Khosla, Aditya and Vondrick, Carl and Torralba, Antonio},
  journal={Advances in neural information processing systems},
  volume={28},
  year={2015}
}

@inproceedings{lian2018believe,
  title={Believe it or not, we know what you are looking at!},
  author={Lian, Dongze and Yu, Zehao and Gao, Shenghua},
  booktitle={Asian Conference on Computer Vision},
  year={2018},
  organization={Springer}
}

@inproceedings{chong2020detecting,
  title={Detecting attended visual targets in video},
  author={Chong, Eunji and Wang, Yongxin and Ruiz, Nataniel and Rehg, James M},
  booktitle=CVPR,
  year={2020}
}

@inproceedings{dias2020gaze,
  title={Gaze estimation for assisted living environments},
  author={Dias, Philipe Ambrozio and Malafronte, Damiano and Medeiros, Henry and Odone, Francesca},
  booktitle={Proceedings of the IEEE/CVF winter conference on applications of computer vision},
  year={2020}
}

@article{cao2019openpose,
  title={Openpose: Realtime multi-person 2d pose estimation using part affinity fields},
  author={Cao, Zhe and Hidalgo, Gines and Simon, Tomas and Wei, Shih-En and Sheikh, Yaser},
  journal={IEEE transactions on pattern analysis and machine intelligence},
  volume={43},
  number={1},
  year={2019},
  publisher={IEEE}
}

@inproceedings{martinez2017simple,
    title        = {A simple yet effective baseline for 3d human pose estimation},
    author       = {Martinez, Julieta and Hossain, Rayat and Romero, Javier and Little, James J},
    year         = 2017,
    booktitle    = ICCV,
}

@inproceedings{zhaoCVPR19semantic,
    title        = {Semantic Graph Convolutional Networks for 3D Human Pose Regression},
    author       = {Zhao, Long and Peng, Xi and Tian, Yu and Kapadia, Mubbasir and Metaxas, Dimitris N.},
    year         = 2019,
    booktitle    = CVPR,
}

@inproceedings{sun2017compositional,
    title        = {Compositional human pose regression},
    author       = {Sun, Xiao and Shang, Jiaxiang and Liang, Shuang and Wei, Yichen},
    year         = 2017,
    booktitle    = ICCV,
}

@inproceedings{yang20183d,
    title        = {3d human pose estimation in the wild by adversarial learning},
    author       = {Yang, Wei and Ouyang, Wanli and Wang, Xiaolong and Ren, Jimmy and Li, Hongsheng and Wang, Xiaogang},
    year         = 2018,
    booktitle    = CVPR,
}

@inproceedings{hossain2018exploiting,
    title        = {Exploiting temporal information for 3d human pose estimation},
    author       = {Hossain, Mir Rayat Imtiaz and Little, James J},
    year         = 2018,
    booktitle    = ECCV,
}

@inproceedings{liu2020comprehensive,
    title        = {A comprehensive study of weight sharing in graph networks for 3d human pose estimation},
    author       = {Liu, Kenkun and Ding, Rongqi and Zou, Zhiming and Wang, Le and Tang, Wei},
    year         = 2020,
    booktitle    = ECCV,
}

@inproceedings{xu2021graph,
    title        = {Graph stacked hourglass networks for 3d human pose estimation},
    author       = {Xu, Tianhan and Takano, Wataru},
    year         = 2021,
    booktitle    = CVPR,
}

@inproceedings{zhao2022graformer,
    title        = {GraFormer: Graph-Oriented Transformer for 3D Pose Estimation},
    author       = {Zhao, Weixi and Wang, Weiqiang and Tian, Yunjie},
    year         = 2022,
    booktitle    = CVPR,
}

@inproceedings{pavllo20193d,
    title        = {3d human pose estimation in video with temporal convolutions and semi-supervised training},
    author       = {Pavllo, Dario and Feichtenhofer, Christoph and Grangier, David and Auli, Michael},
    year         = 2019,
    booktitle    = CVPR,
}

@inproceedings{wang2020motion,
    title        = {Motion guided 3d pose estimation from videos},
    author       = {Wang, Jingbo and Yan, Sijie and Xiong, Yuanjun and Lin, Dahua},
    year         = 2020,
    booktitle    = ECCV,
}

@inproceedings{zhang2022mixste,
    title        = {MixSTE: Seq2seq Mixed Spatio-Temporal Encoder for 3D Human Pose Estimation in Video},
    author       = {Zhang, Jinlu and Tu, Zhigang and Yang, Jianyu and Chen, Yujin and Yuan, Junsong},
    year         = 2022,
    booktitle    = CVPR,
}
}

\end{document}